\documentclass[10pt,twocolumn,letterpaper]{article}

\usepackage[pagenumbers]{cvpr}
\usepackage[accsupp]{axessibility}

\usepackage[dvipsnames]{xcolor}

\definecolor{cvprblue}{rgb}{0.21,0.49,0.74}
\usepackage[pagebackref,breaklinks,colorlinks,citecolor=cvprblue]{hyperref}

\def\paperID{5156}
\def\confName{CVPR}
\def\confYear{2024}

\title{Explaining the Implicit Neural Canvas: Connecting Pixels to Neurons\\ by Tracing their Contributions}

\author{ Namitha Padmanabhan$^{*}$ \quad Matthew Gwilliam$^{*}$ \quad Pulkit Kumar \quad Shishira R Maiya \\ \quad Max Ehrlich \quad Abhinav Shrivastava \\[0.5em]
University of Maryland, College Park \\
{\tt\small ~\url{https://namithap10.github.io/xinc}}
}

\begin{document}
\maketitle
\def\thefootnote{*}\footnotetext{Equal contribution}

\begin{abstract}
    The many variations of Implicit Neural Representations (INRs), where a neural network is trained as a continuous representation of a signal, have tremendous practical utility for downstream tasks including novel view synthesis, video compression, and image super-resolution. Unfortunately, the inner workings of these networks are seriously under-studied. Our work, eXplaining the Implicit Neural Canvas (XINC), is a unified framework for explaining properties of INRs by examining the strength of each neuron's contribution to each output pixel. We call the aggregate of these contribution maps the Implicit Neural Canvas and we use this concept to demonstrate that the INRs we study learn to ``see'' the frames they represent in surprising ways. For example, INRs tend to have highly distributed representations. While lacking high-level object semantics, they have a significant bias for color and edges, and are almost entirely space-agnostic. We arrive at our conclusions by examining how objects are represented across time in video INRs, using clustering to visualize similar neurons across layers and architectures, and show that this is dominated by motion. These insights demonstrate the general usefulness of our analysis framework.
\end{abstract}

\section{Introduction}
\label{sec:intro}

\begin{figure}[h]
  \centering
   \includegraphics[width=\linewidth]{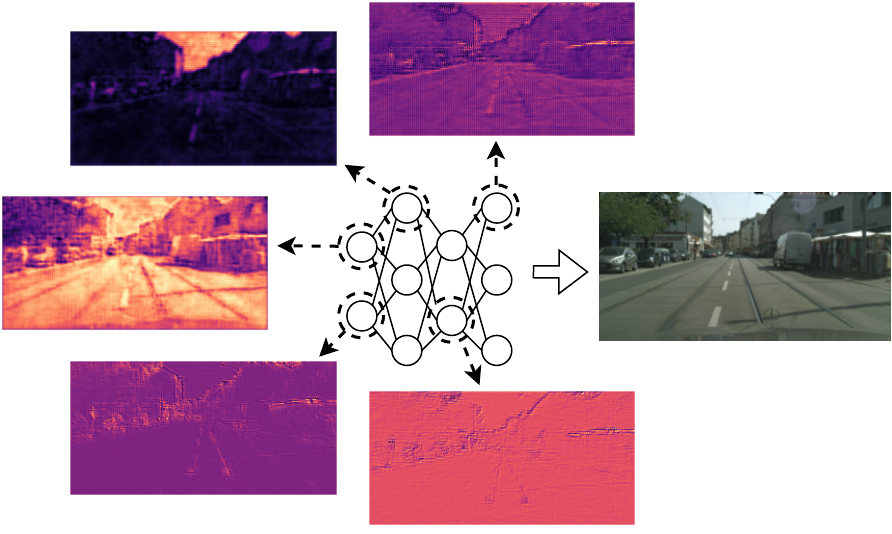}
  \caption{How do INRs ``see'' the images they represent? We propose XINC, which we use to show what parts of a learned visual signal are important to each neuron of an INR. 
  Here, we take a sample from the neural canvas, with contribution maps for 5 neurons sampled from the last layer of a NeRV trained on a Cityscapes~\cite{cordts2016cityscapes} video. Some neurons attend to colors and textures, while others focus on low-level features like edges.}
  \label{fig:teaser}
\end{figure}

Leveraging neural networks to represent data, where the network computes feature maps and embeddings for various visual inputs, is central to computer vision.
Recently, implicit neural representations (INRs) have emerged as an exciting, radically different approach where a multi-layer perceptron (MLP) or convolutional neural network (CNN) is overfit on an image as a generative network, becoming a continuous approximation of the discrete visual data~\cite{saito2019pifu,niemeyer2020differentiable,sitzmann2020implicit,sitzmann2020scene,oechsle2019texture}.
The most well-known type of INRs are Neural Radiance Fields, where a network learns to represent a scene, allowing for continuous scene representation for novel view synthesis, editing, \textit{etc.}~\cite{mildenhall2020nerf,pumarola2021d,zhang2020nerf++,yu2021pixelnerf,gao2023nerf}.
However, INRs are useful for many other tasks as well, in particular, visual data compression~\cite{maiya2023nirvana,chen2021nerv,chen2022cnerv,chen2023hnerv,Zhao_2023_CVPR,He_2023_CVPR,e25081167,dupont2021coin,dupont2022coin++,strumpler2022inrcompress,10.1145/3581783.3613834,zhang2021implicit}.

In spite of their immense utility, the inner workings of INRs remain relatively under-explored.
While some studies shed light on MLP-based INRs~\cite{yüce2022structured}, NeRF~\cite{zhang2020nerf++}, and the impact of hypernetworks~\cite{chen2022transformers,yüce2022structured}, the hidden details of non-coordinate methods such as NeRV~\cite{chen2021nerv} are quite opaque.
This is not completely unique to INRs -- explainability and interpretability are still massive challenges for other areas in vision as well~\cite{gilpin2018explaining,arrieta2020explainable,adadi2018peeking,nauta2023anecdotal}. 
Nevertheless, there is a significant gap for understanding INRs, considering that popular analysis techniques such as NetDissect~\cite{bau2017network} and GradCAM~\cite{selvaraju2017grad} do not transfer to INRs in straightforward ways.
Even analysis work for generative networks tends to focus on control and manipulation~\cite{härkönen2020ganspace}, and the coordinate-to-signal paradigm of INR is distinct enough from the generative noise-to-signal paradigm that most analysis work cannot be trivially extended~\cite{nagisetty2022xaigan}.

While understanding INRs would obviously be beneficial for improving their representation, analyzing with the intent to explain is valuable on its own.
As INRs trend towards commercial viability for tasks like compression, explaining their behaviors becomes critical for real-world enterprises to consider them as alternatives to standard codecs.
Traditional compression methods are not perfect, but their failure modes are known and easily explainable.
Deep INR methods, by contrast, have failure modes that are somewhat unknown and quite opaque.
Thus, beyond helping researchers to improve the models' performance, explaining INRs will help with their actual adoption in the real world.

Our work proposes eXplaining the Implicit Neural Canvas (XINC) as a novel framework for understanding INRs.
Specifically, we propose a technique for dissecting INRs to create contribution maps for each ``neuron'' (group of weights), which connect neurons and individual pixels.
Together, these contribution maps comprise the ``implicit neural canvas'' for a given visual signal.
XINC can thus be used to help us understand the relationships between INRs and the images and videos they represent.

We analyze MLP-based and CNN-based INRs.
While some other analysis works help explain MLP-based INRs~\cite{yüce2022structured}, our analysis of convolution-based INRs is first of its kind. %
To demonstrate the utility of XINC we use it to derive novel insights by characterizing INRs in 5 key ways.

\smallskip
\noindent\textbf{Neuron Contributions Correlate with Colors and Edges} We show representative examples of contribution maps for MLPs and CNNs, and show how they attend to objects in terms of low-level features like edges and colors, while lacking coordinate proximity-based correlations.

\smallskip\noindent\textbf{INRs Must Represent to Omit} We use sums of contribution maps to show that the representations learned by INRs, while intensity-biased, are not dictated by raw pixel magnitude alone. INRs must learn representations even for missing objects, where the final output colors are near zero.

\smallskip
\noindent\textbf{Representation is Distributed} We examine how many pixels each neuron represents, and to what extent, as well as conversely how many neurons are involved in the representation of each pixel. We demonstrate that these representations are quite distributed in nature.

\smallskip
\noindent\textbf{Motion Drives Contribution Change Over Time} We show how NeRVs handle objects and motion, demonstrating that contributions to a given object remain fairly constant over time. This is true even with motion, and fluctuations in our neural canvas across time for videos prove that motion drives changes in neuron contributions for NeRV.

\smallskip
\noindent\textbf{We Can Group Similar Neurons} We cluster neurons from various INRs to show how we can group neurons with similar representation properties using contribution maps.

\section{Related Work}
\label{sec:background}

\noindent\textbf{Implicit Neural Representation} is a way to represent and compactly encode a variety of high resolution signals. The key idea is to map a set of coordinates for a specific signal to a function, by employing a neural network as the continuous approximating function~\cite{sitzmann2020implicit,mildenhall2020nerf,chen2021nerv,xu2022signal,saragadam2023wire}. 
SIREN \cite{sitzmann2020implicit} utilizes periodic activation functions in MLPs to fit and map complicated signals, including images, 3D shapes and videos using very small networks.
COIN \cite{dupont2021coin}, the first image-specific INR method, uses implicit neural representation for image compression tasks and COIN++ \cite{dupont2022coin++} extends this work to encode multiple images. 
In the challenging realm of videos, NeRV \cite{chen2021nerv} is the first method to scale video compression using image-wise implicit representation. 
It uses convolution layers in addition to MLPs and outputs all RGB values of a frame given the positional embedding of frame index $t$ as input. 
Other methods rely on MLPs~\cite{kim2022scalable,maiya2023nirvana}.
Others \cite{chen2022cnerv,chen2023hnerv,He_2023_CVPR,Zhao_2023_CVPR} compute embeddings on the actual frames by introducing a small encoder, thus improving performance over index-based methods.
Meta-learning approaches try to address long per-item fitting times by learning a hypernetwork for predicting network weights~\cite{chen2022transformers,kim2022generalizable}.
Others propose encoding weights as vectors for a wide variety of downstream tasks~\cite{deluigi2023deep}.

\begin{figure*}[t]
  \centering
   \includegraphics[width=\linewidth]{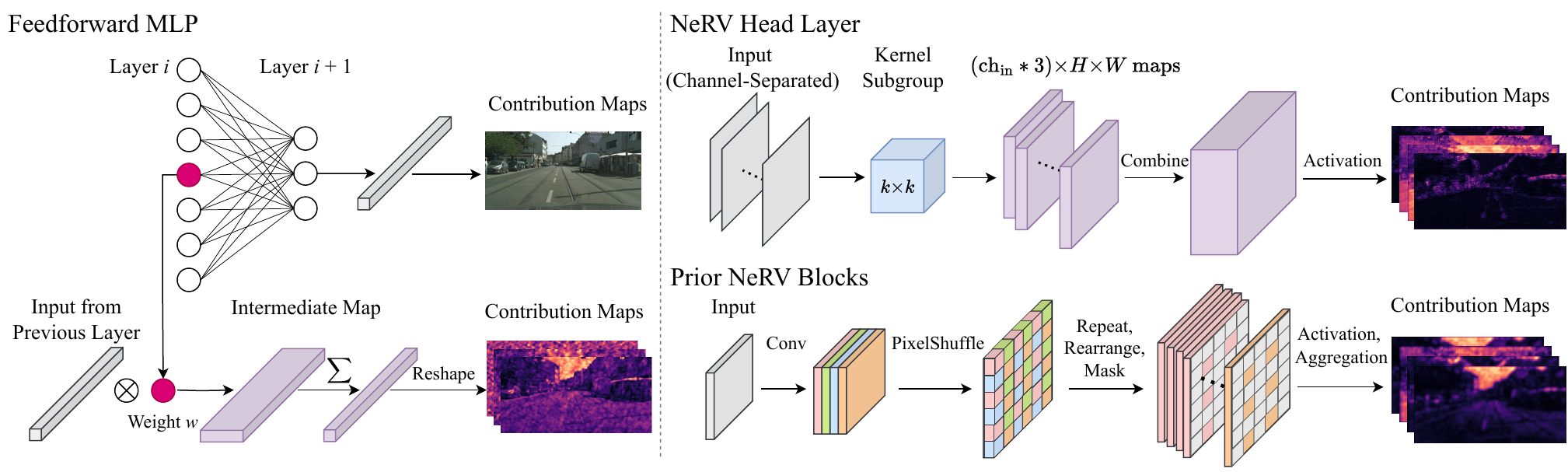}
  \caption{(left) We dissect MLP-based INRs by aggregating their activations (weights multiplied by previous layer outputs) for each pixel at each neuron. (right) We extend this core idea of pixel-to-neuron mapping for the CNN-based INR, NeRV, by computing intermediate feature maps that are not yet summed on the input dimension. For layers prior to the head, we also account for the PixelShuffle and apply an aggregation filter to account for the subsequent layer's kernels, which propagate that neuron's contribution to neighboring pixels. To compute contributions for a given layer, we simply perform the shown steps in sequence for that layer and each subsequent layer.}
  \label{fig:framework}
\end{figure*}

\noindent\textbf{Interpreting and Visualizing Deep Neural Networks} is a longstanding field of study in machine learning.
A large number of techniques aim to understand the internal representations of convolutional neural networks and deep networks in general.
\citeauthor{hoiem2012diagnosing} analyzes effects of various object characteristics on detection performance.
Grad-CAM \cite{selvaraju2017grad} aims to provide visual explanations for the decisions of CNNs.
Using the gradients flowing into the final convolutional layer from a target concept, Grad-CAM and related subsequent works~\cite{draelos2021use,Chattopadhay_2018,fu2020axiombased} generate a coarse localization map that highlights crucial image regions contributing to the prediction of the concept, and \cite{kalibhat2023identifying} proposes a method for automatically captioning these regions to describe image features. 
Network Dissection~\cite{bau2017network} considers each unit in a network as a concept detector and compares their activity with human-interpretable pattern-matching tasks such as the detection of object classes. 
GAN Dissection~\cite{bau2018gan} extends these ideas to visualize and understand GANs at the unit, object, and scene-level. 
Other works characterize neuron behavior with natural language~\cite{hernandez2022natural}, by combining NetDissect-style concepts~\cite{mu2021compositional,wang2022hint,LaRosa2023}, or in terms of how they behave in conjunction with other closely-related neurons~\cite{olah2020zoom}.
\citeauthor{lindsay2023testing} creates a comprehensive test methodology that systematically perturbs model inputs and monitors the effect on model predictions, helping identify potential instabilities. 
These works are insightful for their target domains, but do not directly apply to INR.
So, we propose to analyze the relationships between various parameters in MLP- and CNN-based INR networks and the spatio-temporal locations of the reconstructed output.

\section{Implicit Neural Canvases}
\label{sec:understanding_nerv}

In this section, we explain how we compute the contribution maps that comprise the implicit neural canvas.
These contribution maps connect neurons to pixels in terms of their activations ($\text{weights}{\times}\text{inputs}$) that contribute to each pixel, for INRs. 
We first explain how we do this computation for MLP-based INRs in Section~\ref{subsec:mlp}.
We then explain how we extend this for CNN-based INRs in Section~\ref{subsec:cnn}.

\subsection{Dissecting Multi-Layer Perceptrons}
\label{subsec:mlp}

In this work, as a representative method for MLP-based INRs, we use a Fourier Feature Network (FFN)~\cite{tancik2020fourfeat} with a Random Fourier Matrix~\cite{rahimi2007random} as the positional encoding, followed by linear layers. 
At each MLP layer, the input pixels are operated upon independently, without any spatial rearrangement or sharing of information between the pixels. 
Thus, we can obtain a mapping from the neurons in a layer to the $h{\times}w$ spatial locations (pixels) in the output image.

For each neuron, $\eta$, we compute a mapping of how, given its input, it contributes to every pixel in the final image.
Consider a layer, $l$, with a weight matrix $W$ of shape $m{\times}n$.
We use $i$ and $j$ to denote the weight connecting $i$-th neuron in the previous layer to $j$-th neuron in the current layer, which are also the indices of the current neuron of interest, $\eta_j$, and some neuron in the previous layer, $\eta_i$, respectively.
Now consider the image the INR represents, $I$, of shape $h{\times}w{\times}3$. 
Let us consider that the INR takes pixel location ${x_k, y_k}$ as input, and neuron $\eta_j$ receives an input vector $\textbf{v}$.
We compute its contribution to pixel ${x_k, y_k}$ as $\sum_{i=1}^{m}\textbf{v}_i \cdot w_{ij}$.
When we compute this for all pixels, we obtain a map of shape $h{\times}w$ for $\eta_j$, which we refer to as the contribution map for $\eta_j$.

\subsection{Dissecting Convolutional Neural Networks}
\label{subsec:cnn}
 
For CNNs, we consider kernels as neurons, and we use the NeRV architecture~\cite{chen2021nerv}, which has two types of layers.
Except for the last, all layers consist of a learned convolutional layer, followed by a PixelShuffle~\cite{shi2016realtime} and a nonlinear activation function.
The last layer consists of a convolutional layer and an optional nonlinear activation function.
For this network, the construction of the contribution mapping is less straightforward compared to MLP, since convolutions operate over groups of pixels at each stage, and PixelShuffles rearrange contributions.
We first explain how to address the head layer, and then how to extend this for earlier layers.

\subsubsection{Head Layer}
We begin by analyzing the contribution from kernels in the convolutional head layer to the output RGB image pixels. 
Consider the head layer, $l_h$, with input $v_\text{in}$ of shape $h{\times}w{\times}\text{ch}_\text{in}$ ($\text{ch}_\text{in}$ and $\text{ch}_\text{out}$ are the number of input and output channels, respectively). 
We thus have a set of $\text{ch}_\text{in} * \text{ch}_\text{out}$ convolutional kernels, meaning  $\text{ch}_\text{in}$ kernels per each R, G, B channel of the output, and thus, $\text{ch}_\text{in} * 3$ kernels that can potentially contribute to each output pixel.
We obtain the individual contribution of each kernel by passing each kernel in $l_h$ over $v_\text{in}$ and storing the outputs separately, yielding a set of feature maps, $h{\times}w{\times}(\text{ch}_\text{in} * \text{ch}_\text{out})$. 
To get the true output contribution, and thus the final contribution maps for all  $\text{ch}_\text{in} * 3$ neurons, we pass these feature maps through the $l_h$ activation (\textit{e.g.} \textit{tanh}). 

\subsubsection{Prior NeRV Blocks}

All NeRV layers except the head layer perform some upsampling using PixelShuffle~\cite{shi2016realtime}. 
PixelShuffle upsamples feature maps by moving elements from the channel dimension into the spatial dimension. 
Let the number of input channels of the convolution preceding the PixelShuffle in layer $i$ be $\text{ch}_{i,\text{in}}$ and the number of output channels be $\text{ch}_{i,\text{out}}$. 
For upsampling factor $r$, the number of output channels from the PixelShuffle, which is the same as the number of input channels to the $(i+1)^\text{th}$ layer, is~computed~as 
\begin{equation}
    \text{ch}_{i+1,\text{in}} = \frac{\text{ch}_{i,\text{out}}}{r^2}
    \label{eq:pixelshuffle}
\end{equation}

We require a contribution map of shape $h{\times}w{\times}(\text{ch}_{i,\text{in}} * \text{ch}_{i,\text{out}})$.
To obtain this, we first follow the same approach to the head layer and obtain a mapping of shape $\frac{h}{r}{\times}\frac{w}{r}{\times}\text{ch}_{i,\text{in}}{\times}\text{ch}_{i,\text{out}}$. 
Considering the penultimate NeRV layer for the sake of simplicity, we can apply PixelShuffle to this downsampled contribution map and obtain the map of size $h{\times}w{\times}\text{ch}_{i,\text{in}}{\times}\text{ch}_{(i+1),\text{in}}$. 
However, the contributions of distinct output kernels $\text{ch}_{i,\text{out}}$ are now no longer separated since they are moved into the spatial dimension. 

To resolve this, we repeat the operation $r^2$ times, yielding a new map of size $h{\times}w{\times}\text{ch}_{i,\text{in}}{\times}\text{ch}_{(i+1), \text{in}}{\times}r^2$, which has the same dimensions as a mapping of shape $h{\times}w{\times}\text{ch}_{i,\text{in}}{\times}\text{ch}_{i, \text{out}}$, if we substitute according to Equation~\ref{eq:pixelshuffle}. 
While this map has the dimensions we require for further processing by downstream layers, we must take care that each non-overlapping $r{\times}r$ block contains samples from the same filter in order to preserve each filter's contribution. 
If nearby samples were from different filters, as in the traditional PixelShuffle, then the contribution of different filters would be mixed and aggregated in future layers. 
We can conveniently accomplish this with a simple re-arranging of the repeated channels followed by masking such that each $r{\times}r$ block contains only one non-zero element (to prevent overcounting of the contribution of each kernel). 
The final result is then passed through the activation layer before proximity correction. %

Layers before the penultimate layer differ only slightly.
For the layer in question, we treat it exactly like the penultimate layer. 
However, the result is not at the target output resolution.
So, we must perform the subsequent layers' corresponding operations, with one minor adjustment -- we use nearest neighbor upsampling, instead of PixelShuffle, for all layers after the initial layer.
This is because the spatial correspondences of the PixelShuffle are specific to the feature map structure of the given subsequent layer, and have no meaning for the earlier layer feature maps.

\begin{table}[t]
\begin{minipage}{\linewidth}
\begin{center}
\caption{\textbf{Dataset Statistics.} For these densely annotated datasets we provide the typical number of annotated instances per video, video length, and the portion of frames which are annotated.}
\label{tab:dataset-statistics}
\vspace{0.25em}
    \resizebox{\textwidth}{!}{
    \small\begin{tabular}{@{}l| c c c c c@{}} 
        \toprule
        Dataset &  Videos & Instances  & Frames & \% Labeled & Domain \\
        \midrule
       Cityscapes-VPS~\cite{kim2020vps} & $500$ & $2-71$ & $30$  & $1/6^\text{th}$ & Streets \\ 
       VIPSeg~\cite{miao2021vspw} &  $3536$ & $1-78$ & $4-81$ & All & Open \\ 
        \bottomrule
    \end{tabular}
}
\end{center}
\end{minipage}  
\end{table}

\subsubsection{Proximity Correction}

Layers before the head layer do not contribute directly to pixels. 
Instead, their outputs are the inputs to some later set of convolution kernels.
While we account for the upsampling with how we handle PixelShuffle, we still must address the fact that the subsequent kernels act on $k{\times}k$ neighborhoods at each filter map.
So, we use an aggregation filter to account for the fact that a given point in some intermediate feature map contributes to a $k{\times}k$ neighborhood of points in the next layer's feature maps. %
This is a simple $k{\times}k$ box filter for each layer.

\section{Understanding Representations}
\label{sec:analysis}

\subsection{Dataset and Training Details}
\label{subsec:preliminaries}

\begin{figure*}[ht]
  \centering
  \includegraphics[width=2\columnwidth]{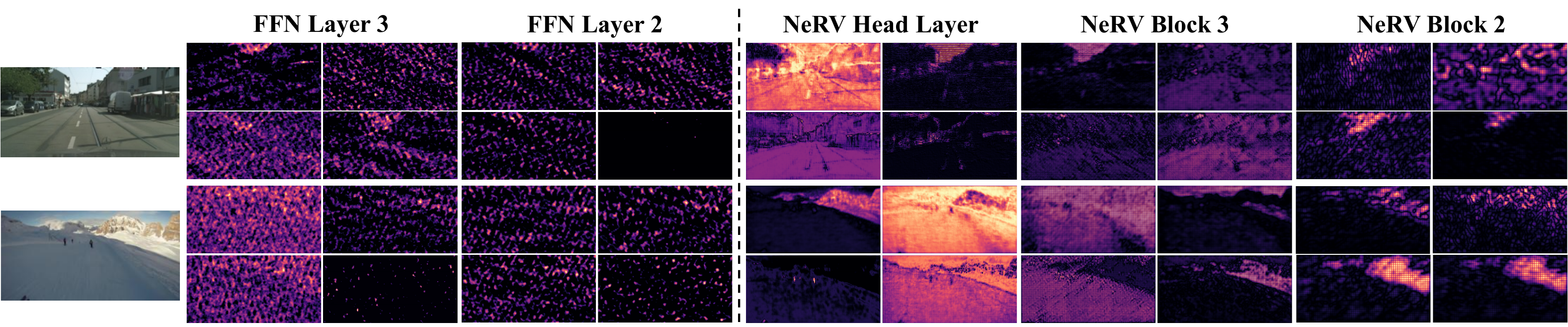}
  \caption{\textbf{The implicit neural canvas.} We show representative example contribution maps for various layers of FFN~\cite{tancik2020fourfeat} and NeRV~\cite{chen2021nerv}. Notice how early layer FFN neurons manifest strong Fourier patterns, and how the last layers of NeRV tend to resemble the image, with NeRV head layer neurons being reminiscent of classical image processing filters.}
  \label{fig:contribution_main}
\end{figure*}

\begin{figure*}
  \centering
   \includegraphics[width=\linewidth]{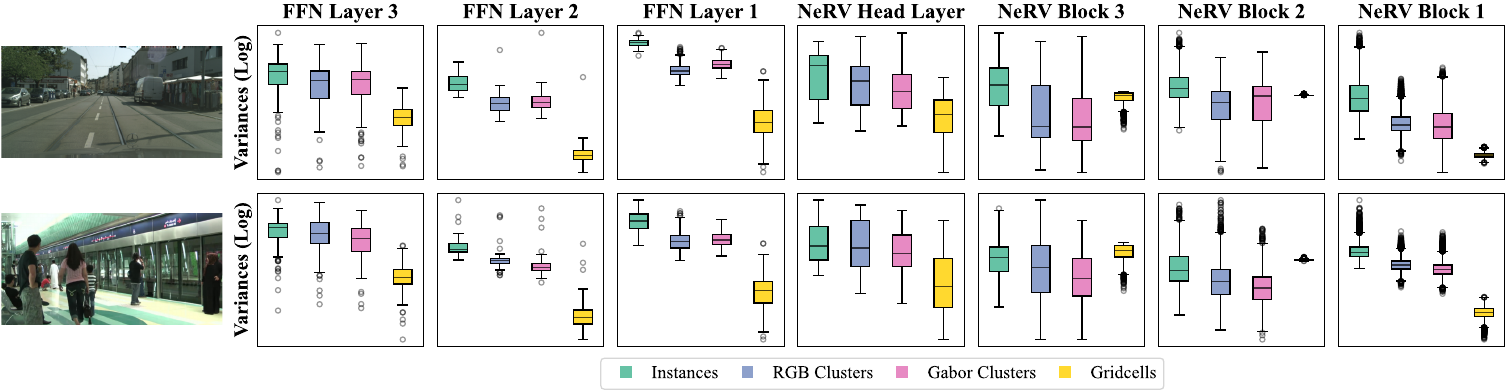}
  \caption{\textbf{Grouping contributions.} We compute the variance of the difference from expected contribution for different groupings of contributions - Instances of objects and background, RGB color-based clusters, Gabor filter-based clusters and regular gridcells. These results suggest that INRs ignore space while preferring instances, color, and edge features.}
  \label{fig:variance_of_patch_deltas}
\end{figure*}

To analyze the relationship between neurons and pixels in encoded signals, we use videos from the Cityscapes-VPS~\cite{kim2020vps} (train and val) and VIPSeg~\cite{miao2021vspw} video panoptic segmentation datasets. 
Table \ref{tab:dataset-statistics} provides the number of instances per frame in each dataset, the percentage of annotated frames and other information.  
To select a few videos from these datasets, we look for examples containing objects of diverse shapes, sizes and categories, to afford diversity in analysis. 
Having instance and object category level information enables us to answer questions pertaining to whether INRs understand semantics and instances and also helps analyze the distribution of representations for objects of different types. 
Further, it would help us determine whether the groups of parameters attending to the same objects remains consistent across frames, whether the model omits representing small objects, and investigate other important properties.
We train our FFNs and NeRVs on downsampled images, to keep the model and contribution map sizes small.
Specifically, we resize and crop all videos to $128{\times}256$ resolution. 
We choose model sizes and training times such that bits-per-pixel and reconstruction quality are roughly equivalent across models and video frames.
We provide more details in the appendix.

\subsection{Contributions Correlate with Colors and Edges}

Figure~\ref{fig:contribution_main} shows some example contribution maps for neurons at different layers for both MLP-based and CNN-based INRs.
FFN and NeRV head layer neurons seem to learn some form of whole scene representation, and the CNN penultimate layer has similar characteristics.
The contribution maps of the FFN layers suggest a progression from representing Fourier features in early layers, to representing more of the image in question, with Fourier artifacts, in the last layer.
For the NeRV layers, there seem to be a variety of features captured by various neurons, including edges, textures, colors, and depth.
The behavior of a neuron, in both networks, is not simply to represent some pixels, instead, it learns some low-level scene attributes.

\begin{figure*}
    \includegraphics[width=\linewidth]{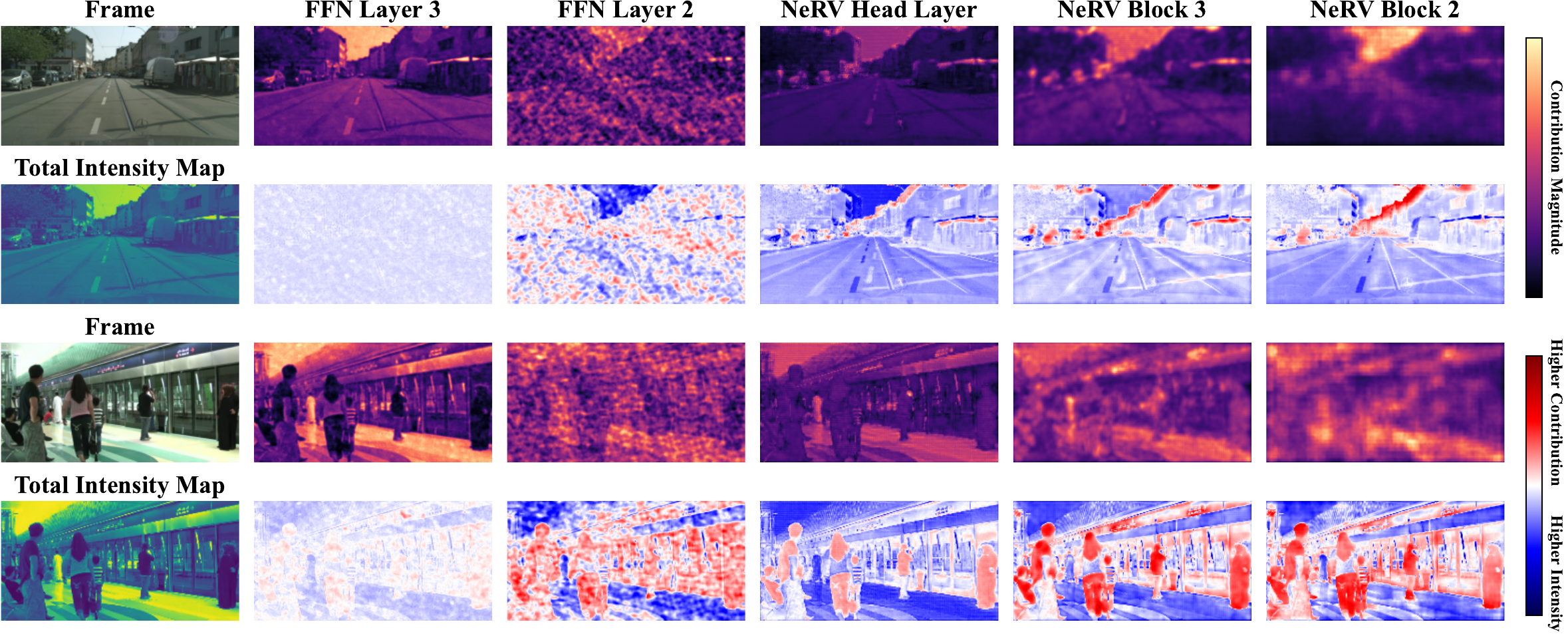}
  \caption{\textbf{Contribution vs. intensity.} We compare contribution and intensity in alternating rows. In the top row, we sum all contribution maps for the indicated layer. In the next row, we show the difference between this, and the raw image intensity (sum of all color channels) to show when contribution does and does not correlate with intensity. The next two rows repeat this for a frame from another video.}%
  \label{fig:contribution_vs_intensity_map}
\end{figure*}

\begin{figure}[ht]
  \centering
   \includegraphics[width=\linewidth]{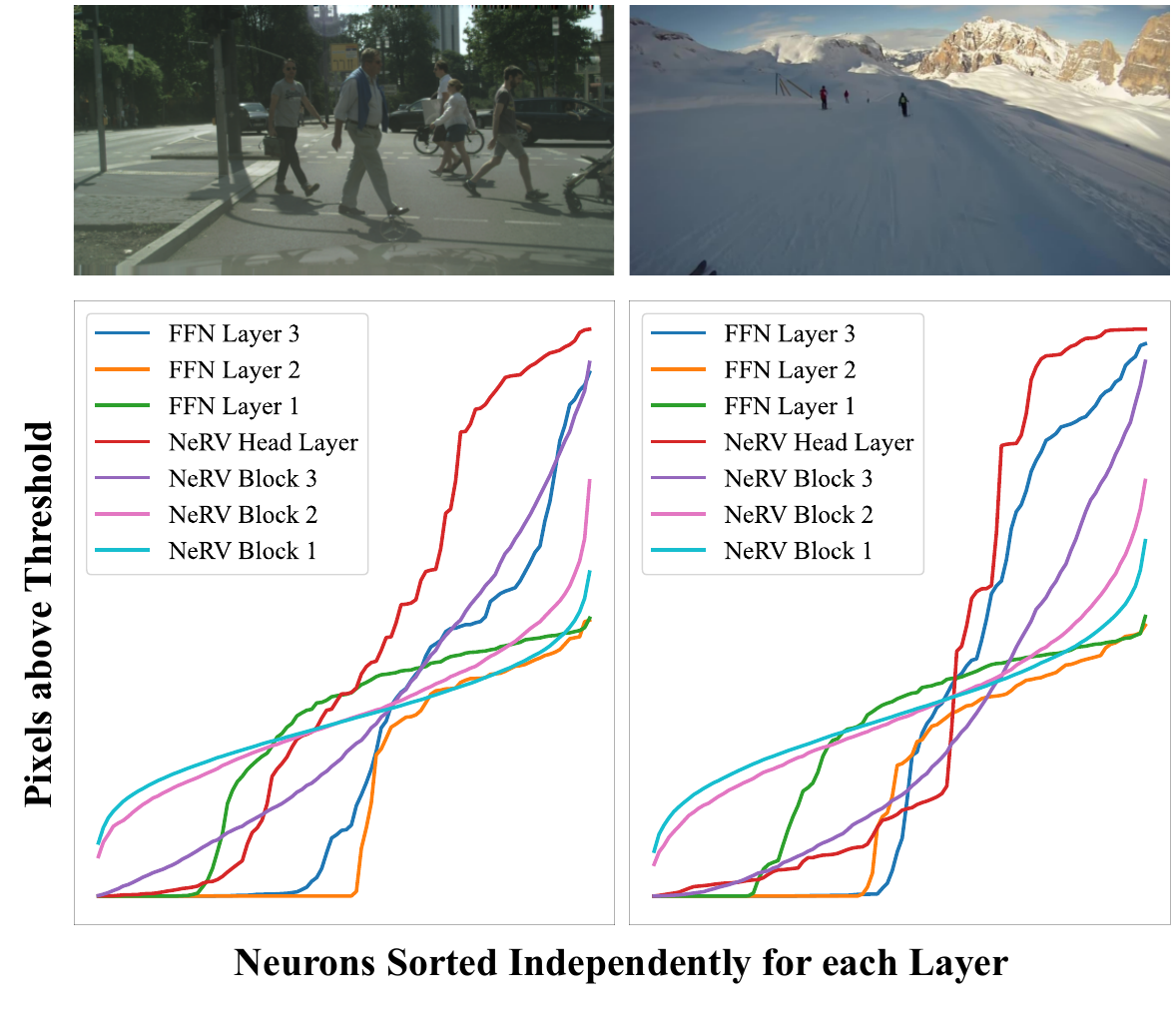}
  \caption{\textbf{Pixels per neuron.} We compute the pixels activated for each neuron in each layer, with each layer sorted independently by order of increasing number of pixels, for frames from two videos.}%
  \label{fig:pixels_per_neuron}
\end{figure}

We explore these behaviors quantitatively by aggregating neuron contributions to pixels, first in terms of instances, and then also by clustering pixels by color value (RGB Clusters), clustering on Gabor filter features~\cite{/content/journals/10.1049/ji-3-2.1946.0074,MEHROTRA19921479} (Gabor Clusters, details in appendix), and clustering pixels spatially (taking equal size rectangular gridcells).
We hypothesize that for a type of pixel clustering to be meaningful for neurons, each neuron should tend to have high activation in some clusters, and little to no activation in others. 
Quantitatively, the variance of contributions across clusters would be high.
However, using raw variance of contributions would give massive bias to clusterings that form large and small regions, where some contributions would be trivially small.
So instead, we first compute the expected contribution for each cluster, which is the contribution over the whole image, normalized by size of the given cluster. 

\begin{equation}
    \small \text{cont}_\text{expected} = \left(\frac{\text{area}_\text{cluster}}{\text{area}_\text{image}}\right) * \text{cont}_\text{image}
\end{equation}
We then obtain the actual contributions to each cluster, $\text{cont}_\text{actual}$ by summing the contributions of a neuron to the pixel values in the cluster,
and contribution deltas are taken as $\delta_\text{cont} = \text{cont}_\text{expected} - \text{cont}_\text{actual}$.
We then normalize each $\delta_\text{cont}$ by dividing by $\text{cont}_\text{expected}$ such that they are percentage differences between actual and expected neuron contributions to each cluster.
Finally, we compute the variance (standard deviation) of the normalized percentage differences. %
The above is repeated for all neurons in a layer.

We plot the result of performing this computation for instances, RGB, Gabor, and gridcell clusters, across all neurons, in Figure~\ref{fig:variance_of_patch_deltas}.
This computation reveals quantitatively what Figure~\ref{fig:contribution_main} hints at -- there is relatively little meaningful correlation between contributions and space, as seen in the low variances of gridcell representations.
Instead there is in general a surprisingly high variance for neurons when we group their contributions using instance masks, which suggests that they learn some type of object semantics.
Given that the variances for RGB and Gabor clusters are also quite high, we suggest that these are low-level semantics, a combination of color and edge features.
There is also an interesting trend with the NeRV Block 3, where it seems to have some meaningful correlation with gridcells.
However, we note that the top 25\% of variances for Instances, RGB, and Gabor, are almost all higher than even the highest gridcell variance, and therefore we believe the intuition holds.

\subsection{Contribution is not Intensity}

\begin{figure*}[t]
  \centering
   \includegraphics[width=\linewidth]{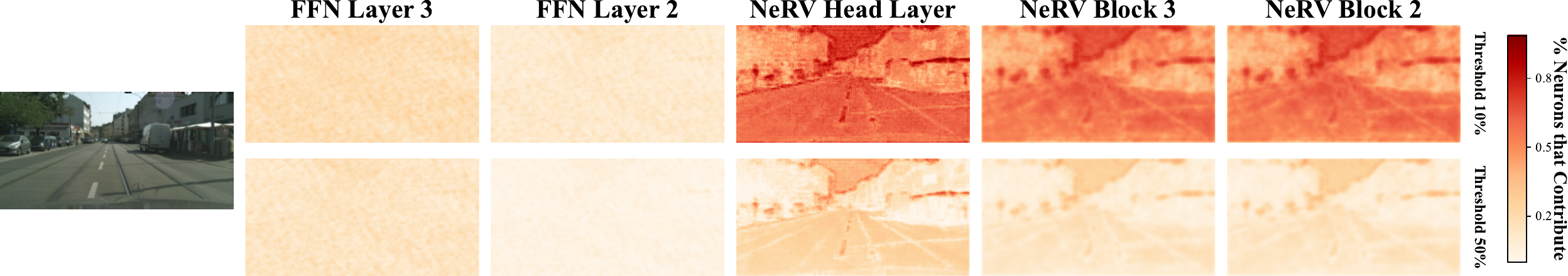}
  \caption{\textbf{Neurons per pixel.} We show how many neurons represent significant portions of each pixel, as a percentage of the total neurons in the indicated layer, at two different thresholds for ``activation.''}
  \label{fig:neurons_per_pixel}
\end{figure*}

\begin{figure*}[t]
  \centering
   \includegraphics[width=\linewidth]{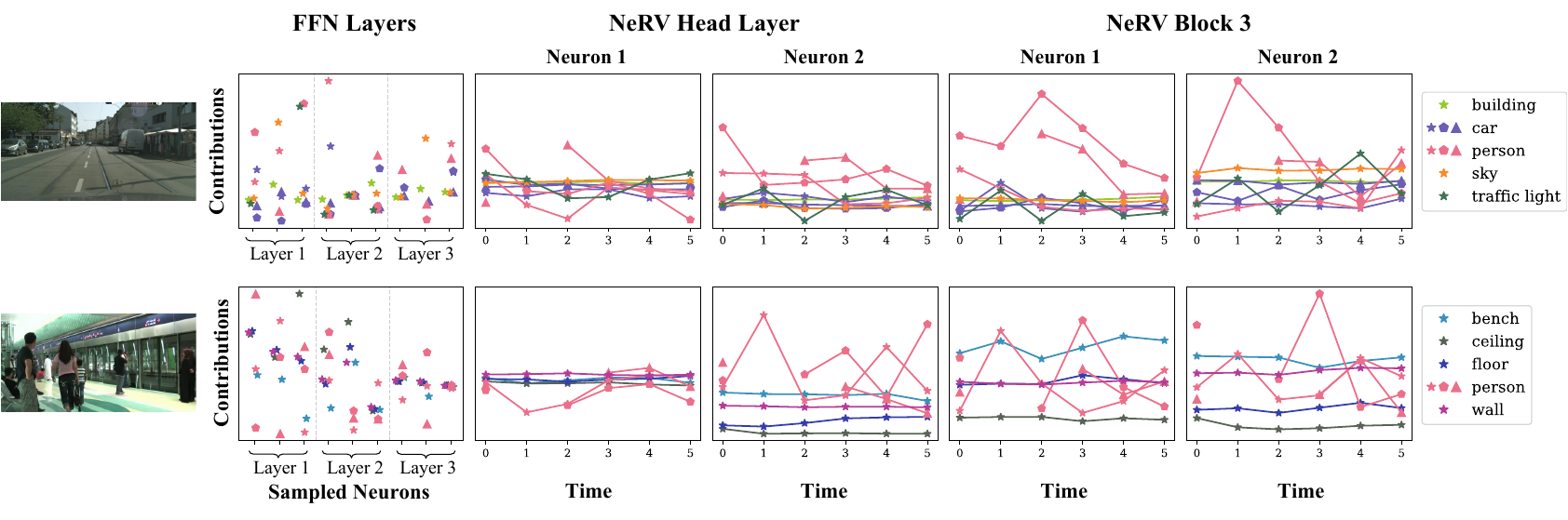}
   \vspace{-2.0em}
  \caption{\textbf{Neuron contributions to things and stuff.} We aggregate the contributions of neurons to the instances (things) and background (stuff) in a frame from 2 videos. For FFN, we sample 3 neurons from each layer and for NeRV, we sample 2 neurons each from the last 2 layers. Instances belonging to the same category are depicted by distinct markers of the same color. For FFNs, we show contributions only for the given frames, whereas for NeRVs, we show these contributions over time for 6 frames from the source video.}
  \label{fig:kernel_to_instance_contribs}
\end{figure*}

Furthermore, we observe an interesting property that not all pixels require equal representation.
Figure \ref{fig:contribution_vs_intensity_map} shows the sum of absolute contributions over all neurons for distinct layers of the FFN and NeRV. 
We compare these aggregated, layer-level contribution maps to the total intensity (summing over color channels) of the image they correspond to. 
The heatmaps reveal a correspondence between the areas of large contributions and the areas of high intensity in the input image. 
This suggests that the neurons of an INR largely attend to higher intensities in the image.

While there is naturally some intensity bias, as a result of the pixelwise loss functions used by these networks, this does not fully explain the behavior of the representation.
To discover other concepts that the INR potentially pays attention to, we subtract the intensity image from the contribution heat map. 
In Figure~\ref{fig:contribution_vs_intensity_map}, this reveals that in addition to intensity, the INR layers also pay attention to other low-level information such as edges and textures.
We note that the NeRV layers and the FFN layer 2 have a tendency to over-represent the people in the second image, relative to intensity, while the FFN layer 3 contributions are largely consistent with intensity.
Additionally, not all edges are treated equally, with NeRV having large contributions for some edges, and small contributions for others.

\subsection{Representation is Distributed}
\label{subsec:representation_distributed}

When training an INR, one might wonder, since there are more pixels than neurons, how many pixels are represented by each neuron, and if this representation tends to be distributed. 
To analyze the manner in which neuron-to-pixel contributions are distributed, we compute the ratio of a neuron's contribution to a pixel to the total contribution across neurons at a pixel. 
We threshold on this ratio to determine whether a pixel receives a significant portion of its contribution from a specific neuron or not. The threshold used is a simple reciprocal of the number of neurons in a layer.
We then compute the total number of pixels that a specific neuron contributes to meaningfully and sort this across neurons, repeating this analysis individually for different layers of the FFN and NeRV. 

We show these contribution counts in Figure~\ref{fig:pixels_per_neuron}. %
Since different layers have different numbers of neurons, we resample the neurons of each layer to a common length using linear interpolation to preserve their trends. Notice how the outermost layers of the FFN and NeRV (layer 3 and head layer respectively) show the highest trends with a decreasing trend towards the earlier layers of both networks. 
Note that nearly $50\%$ of the neurons in FFN Layer 2 have almost no contributions, producing a flat line in the first half.
The presence of these dead neurons in this and other layers points to the suitability of these networks for data compression, and helps explain why extensive model pruning can be done with limited impact on reconstruction quality~\cite{chen2021nerv}.

To further understand how the learned INRs distribute the image representations, we compute at each spatial image location, the density of neurons that contribute to it above a threshold, $\tau$. 
To set $\tau$ for a certain layer, consider the sorted raw contributions of all neurons across locations of the image. 
We find that a large portion of the contributions are small and the number of larger contributions is fewer in comparison.
For an illustration of this, see Figure~\ref{fig:distribution_of_contributions}. 
We obtain the raw contribution value that lies at the $x$-th percentile of this curve, i.e., the contribution of the neuron to the left of which lies $x\%$ of the area under the curve. 
The value corresponding to this percentile is chosen as the threshold for all neuron contribution maps. 
We show this for the $10^\text{th}$ and $50^\text{th}$ percentile thresholds in Figure \ref{fig:neurons_per_pixel}, which shows the regions of the image with a higher density of dedicated neurons. 

High intensity and highly textured areas not only have high contributions as seen earlier in Figure \ref{fig:contribution_vs_intensity_map}, but they also have a large portion of neurons dedicated to them. 
Increasing the threshold highlights the image areas with the highest neuron density, and this reveals an overwhelming bias for edges with NeRV.
However, all these trends are greatly subdued for the FFN at both its head and penultimate layers, suggesting that while the magnitude of its contributions 
correlates almost perfectly with the image intensity, its actual representation is more evenly distributed by comparison.

\subsection{Objects, Categories, and Motion}

To analyze whether INRs have meaningful representations in terms of objects, we aggregate neuron contributions over pixels in each instance's segmentation map. 
Consider the two sample videos shown in Figure \ref{fig:kernel_to_instance_contribs}. 
Each has a different set of objects with some categories having multiple instances. 
Instances may move, disappear and reappear over time. For the FFN, we sample three neurons each from the three layers and depict each neuron's normalized contributions to every instance (as a percentage of contributions to all instances) in the first frame of the video. 
Notice how contributions to instances of the same category are not grouped together. 
This is related to our findings in Figure~\ref{fig:variance_of_patch_deltas}, and motivates our claim that these networks learn low-level object features, but lack class-level semantics. %

For NeRV, we sample two neurons each from the head layer Block 3, and track the normalized per-instance contributions over multiple frames of the video. 
We see that a neuron's contributions to an instance remains relatively static over time. 
When there is fluctuation, we notice that this is often correlated with instances that move, such as persons. 
Another thing to note is how contributions to high intensity instances such as sky are higher, which is in agreement with our earlier observations.

\begin{figure}
  \centering
   \includegraphics[width=0.95\linewidth]{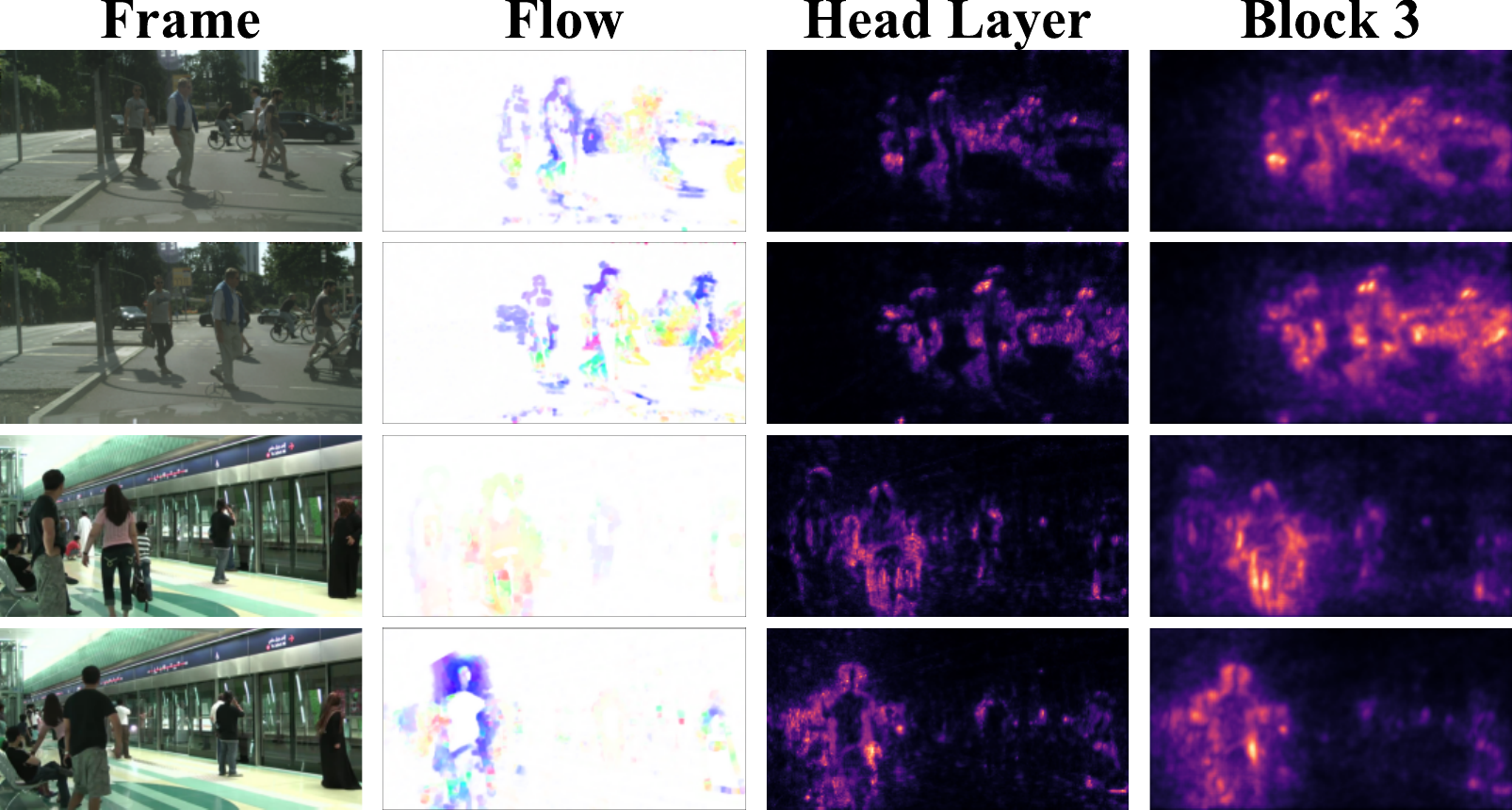}
  \caption{\textbf{Neurons and motion.} We show the correlation between motion and changes in neuron contribution over time by computing optical flow between two frames and the difference in contribution maps between those same frames. We plot both of these. Fluctuation is driven by motion, and it seems the areas revealed by motion matter equally to the objects that are actually moving.}
  \label{fig:optical_flow}
\end{figure}

Further delving into how neurons in NeRV respond to motion, we visualize the optical flow map and the fluctuation of total contribution in the last 2 layers of a NeRV. 
Figure~\ref{fig:optical_flow} depicts this for two different time steps in each video (the flow and contribution fluctuation are computed between a specific frame and the immediately preceding frame). 
We see that the changes in contributions are focused more in areas around the regions of flow. 
It appears that when a moving object reveals new background, this causes fluctuations in the spatially proximal pixels, even though they technically contain no motion themselves.
Thus, we see that in spite of its lack of high-level object semantics, the changes in how a NeRV represents a video across time are dominated by the motion of the entities in the video.

\subsection{Neurons can be Clustered by Contributions}
\label{subsec:clusters}

  \vspace{-0.75em}

\begin{figure}[ht]
  \centering  \includegraphics[width=0.95\linewidth]{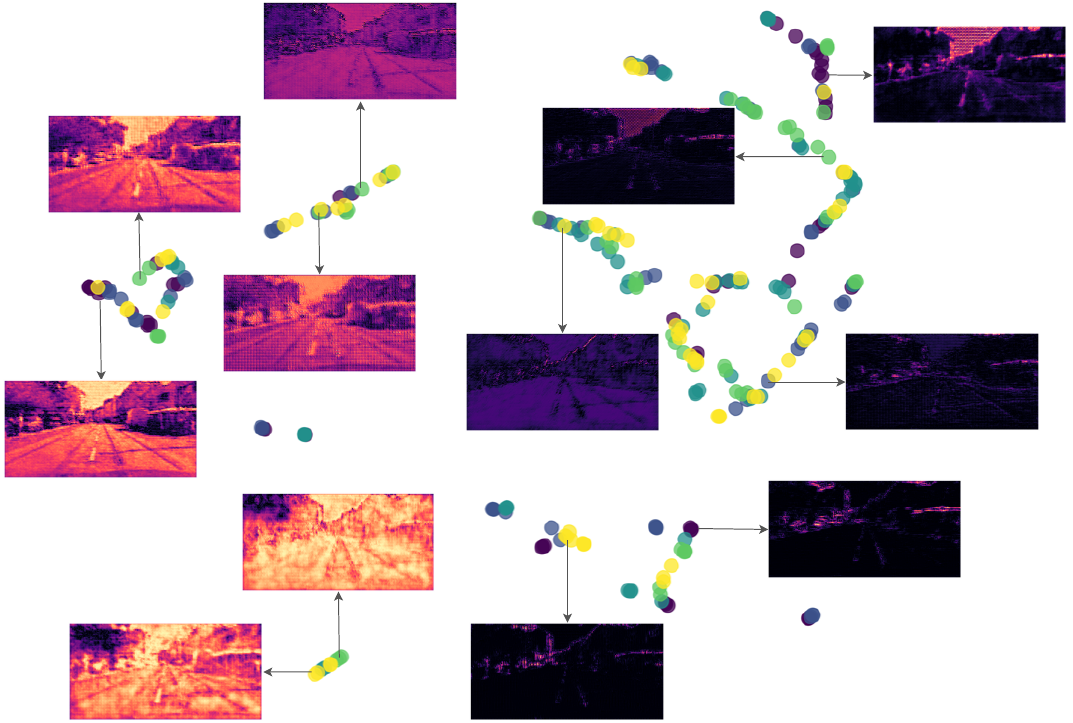}
  \caption{\textbf{UMAP for INRs.} We plot the neurons from the head layers of NeRVs with different seeds, and show the contribution maps corresponding to the neurons indicated.}
  \label{fig:inr_umap}
\end{figure}

We propose a unique approach for computing a vector representation of INR neurons. 
Inspired by the relatively high variance for Gabor clusters in Figure~\ref{fig:variance_of_patch_deltas}, we take the pixels in each frame of each video, and cluster them according to Gabor filter features with $k$ clusters.
We then aggregate the contribution map values according to the clusters, yielding a $k$ dimensional vector representation of every neuron.
We can then project these representations with UMAP to allow for plotting them in 2D space, which we do in Figure~\ref{fig:inr_umap}.
As this figure shows, there are many different ``types'' of neurons when we separate them according to edge contributions.
This representation mechanism we develop with XINC provides another powerful method for visualizing and examining the internals of INR.

\section{Conclusion}
\label{sec:conclusion}

In this paper, we propose XINC, a novel framework for analyzing INRs for image and video representation.
Using XINC, we find interesting trends with CNNs and MLPs, such as how while they lack high-level object semantics, they do have biases that mimic object representation, particularly with edges and colors.
We also show how fluctuations in contribution for NeRV are driven by motion.
We show how our framework can help us identify interesting groups of related neurons. 
We provide both the framework and these findings to help drive future research for INRs.

\vspace{1.0em}
\noindent\textbf{Acknowledgements.} We would like to thank Soumik Mukhopadhyay for his helpful feedback while we prepared the manuscript. This project was partially funded by NSF CAREER Award (\#2238769) to AS.

{
    \small
    \bibliographystyle{ieeenat_fullname}
    \bibliography{main}
}

\clearpage

\maketitlesupplementary

The main purpose of this appendix is to expand our results from the main paper by showing analysis for frames from additional videos, to demonstrate that our analysis and findings hold across a variety of different videos.

\section{Implementation Details}

\subsection{INR Architecture Settings}

Neither the INR or NeRV have established out-of-the-box settings for arbitrary, in-the-wild videos.
Thus, we have to carefully explore and select settings that allow for meaningful comparisons.
We try to ensure the FFN and NeRV have similar compression ratios; however, this is not fully practical for one key reason.
NeRV is able to leverage the temporal redundancies in addition to the spatial redundancies.
We also try to ensure that both networks achieve similar reconstruction quality.
For the videos we choose, the NeRVs have a mean PSNR of 33.50, and for the frames, the FFNs have a mean PSNR of 34.29.

For all the NeRVs, we ensure they have 978,557 parameters; for all FFNs, we ensure they have 32,971; for a 30 frame video (for Cityscapes-VPS all are 30 frames, for VIPSeg we only use videos between 30 and 45 frames), where the NeRV represents all 30 frames, and a given FFN represents a single frame, this gives roughly equivalent bits-per-pixel.
For the FFN, we feed the 208-dimension Fourier position encoding to a network with 3 layers, with hidden sizes of 104 for the inner layers, ReLU activations, and output as 3-channel rgb predictions.
For the NeRV, we use 4 layers, with the upsampling layers having strides 4, 2, and 2, respectively. 
We borrow the parameter reduction trick from HNeRV to balance parameters between layers, and set this to $r = 1.2$~\cite{chen2023hnerv}, with a minimum width of 6, and the model size hyperparameter set to 1 million.
The inner layers use GELU activations~\cite{hendrycks2023gaussian}.
For other settings, we use the original NeRV defaults~\cite{chen2021nerv}.
Both are trained for 1000 epochs with L2 loss. 

\subsection{Gabor Features}

We already have mechanisms for addressing some types of low-level and high-level features.
Since we have dense instance and background segmentation masks, we are able to identify how neuron contributions correspond with instances and background.
We can also cluster pixels by color and space.
However, we notice low-level patterns in the contribution maps, such as a tendency to focus on edges, that are not captured fully by instance masks, color, or space.
So, we use an additional mechanism for clustering pixel locations -- Gabor filters~\cite{/content/journals/10.1049/ji-3-2.1946.0074,MEHROTRA19921479}.

Gabor filters offer a robust method for texture analysis, detecting patterns across various orientations and scales. 
In our experiments, we utilize Gabor filters with four orientations and three scales, ensuring a comprehensive analysis of diverse textures within the contribution maps. 
The filters are applied to the maps, producing a distinct feature map for each filter. These feature maps are then stacked. 
For each pixel coordinate, a feature vector is constructed from values across all filters. 
These vectors are utilized to group contributions with similar traits into Gabor clusters (see Section~\ref{subsec:clusters}). 
Please refer to the code for precise settings and implementation.

\section{Further Results}

\subsection{Contribution Maps}

\begin{figure}
  \centering
   \includegraphics[width=0.95\linewidth]{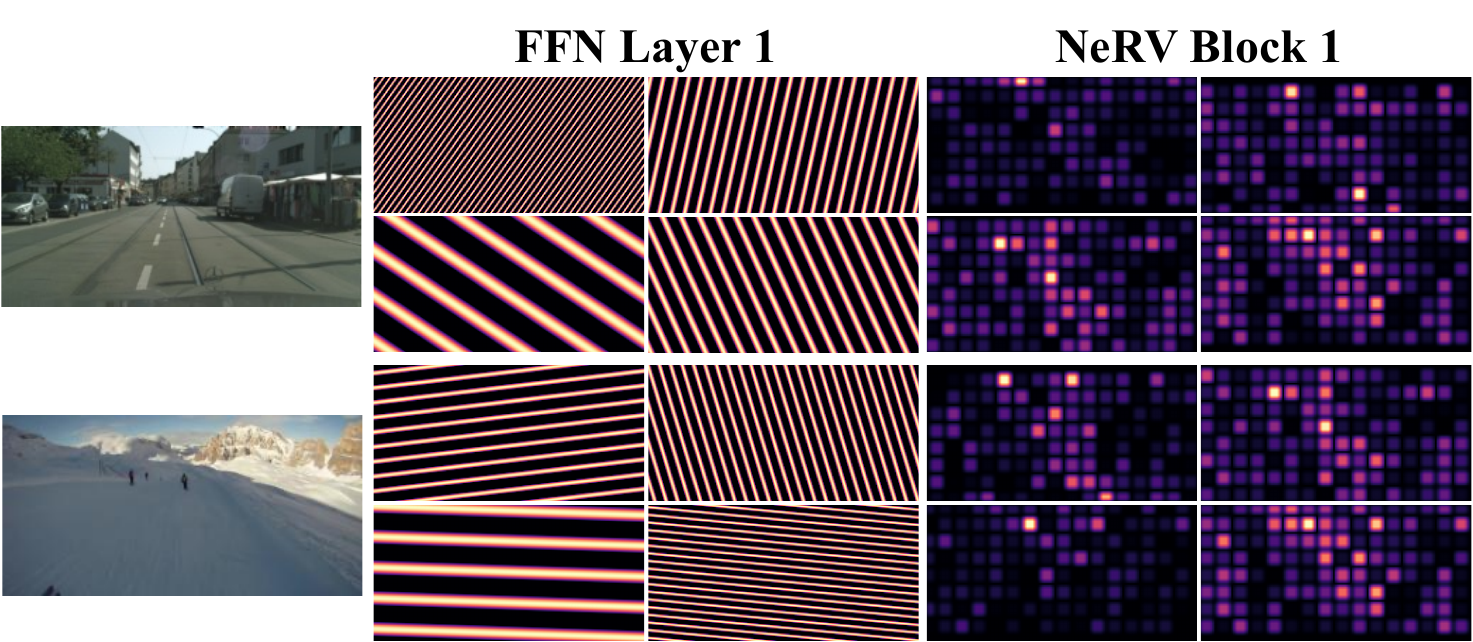}
  \caption{\textbf{The implicit neural canvas.} We show the contribution maps for sample first layer neurons of FFN and NeRV.}
  \label{fig:contribution_supp}
\end{figure}

We show the missing layers from Figure~\ref{fig:contribution_main} in Figure~\ref{fig:contribution_supp}.
Note that the earliest layers have patterns that heavily correlate with the fourier features (FFN) and positional encoding (NeRV). Also, due to the nature of the PixelShuffle, for the first layer of NeRV, some neurons cannot possibly represent certain pixels. 
In that sense, the representation of each neuron is forced to be somewhat sparse.

\begin{figure*}
  \centering
   \includegraphics[width=\linewidth]{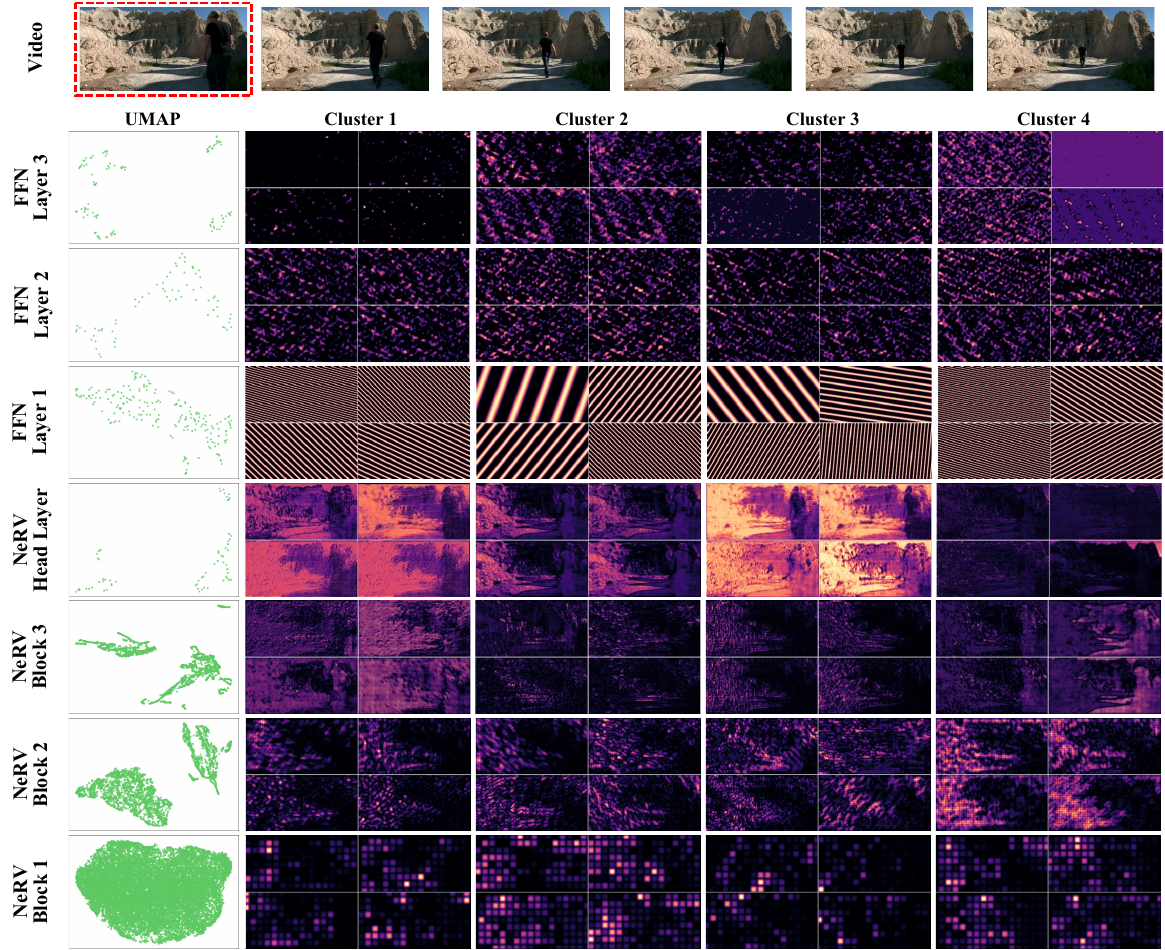}
  \caption{\textbf{Neuron Clusters.} We cluster neurons for each layer/block into 4 clusters, and then sample 4 neurons from each cluster. We show contribution maps of sample clustered neurons for the first (red-bordered) frame for the indicated video (top). See another video in Figure~\ref{fig:vid2_umap_heatmaps}.}
  \label{fig:vid1_umap_heatmaps}
\end{figure*}

\begin{figure*}
  \centering
   \includegraphics[width=\linewidth]{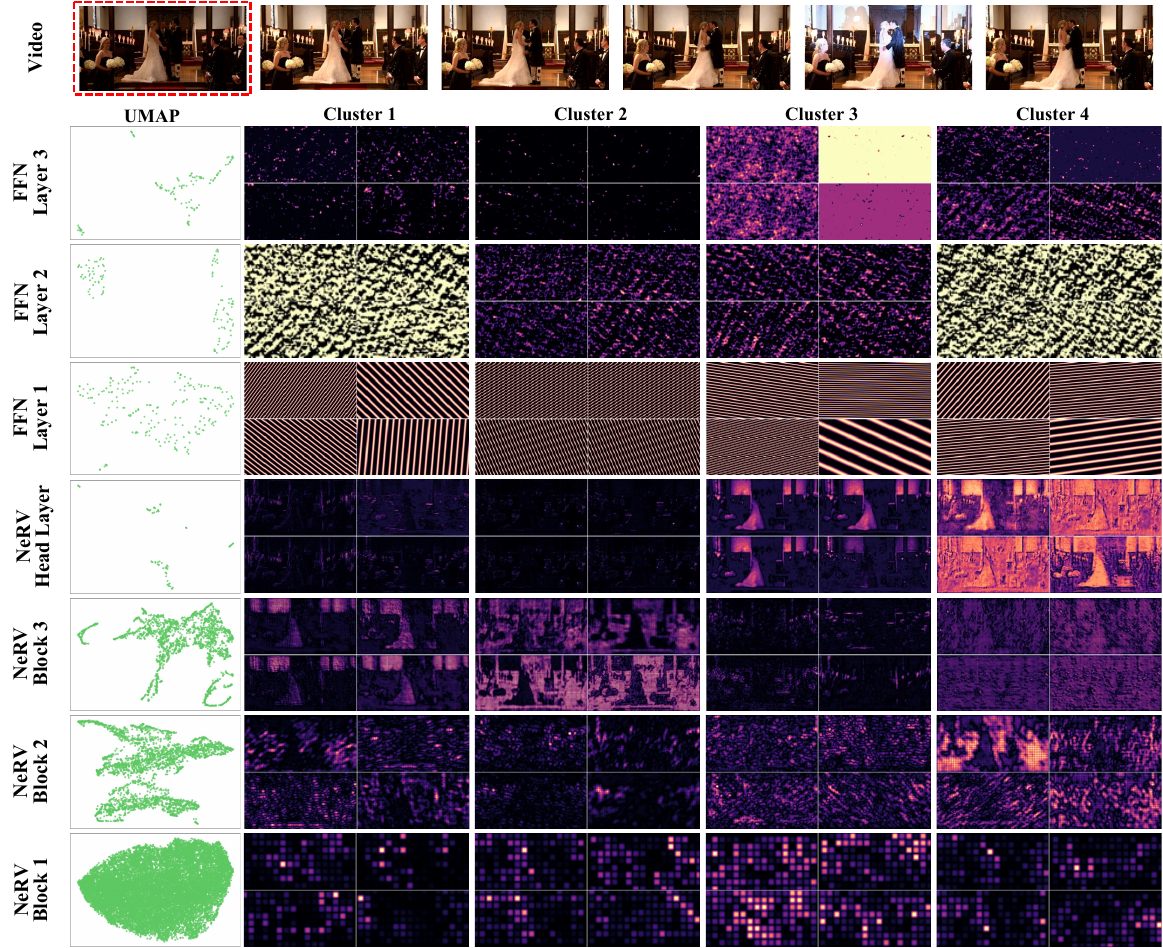}
  \caption{\textbf{Neuron Clusters.} Following the same procedure as in Figure~\ref{fig:vid1_umap_heatmaps}, we cluster neurons for each layer/block into 4 clusters, and then sample 4 neurons from each cluster. We show contribution maps of sample clustered neurons for the first (red-bordered) frame for the indicated video (top).}
  \label{fig:vid2_umap_heatmaps}
\end{figure*}

Using Gabor filter features, we cluster neurons at each layer of MLP-based and CNN-based INRs and plot them using UMAP. For a few of these clusters, Figures \ref{fig:vid1_umap_heatmaps} and \ref{fig:vid2_umap_heatmaps} show the contribution maps of four neurons sampled from each cluster. We see that early FFN layers learn Fourier patterns, and the last layer of NeRV tends to resemble the image. Further, note how neurons belonging to the same cluster tend to learn similar contribution maps as compared to neurons belonging to different clusters.

\begin{figure*}
  \centering
   \includegraphics[width=\linewidth]{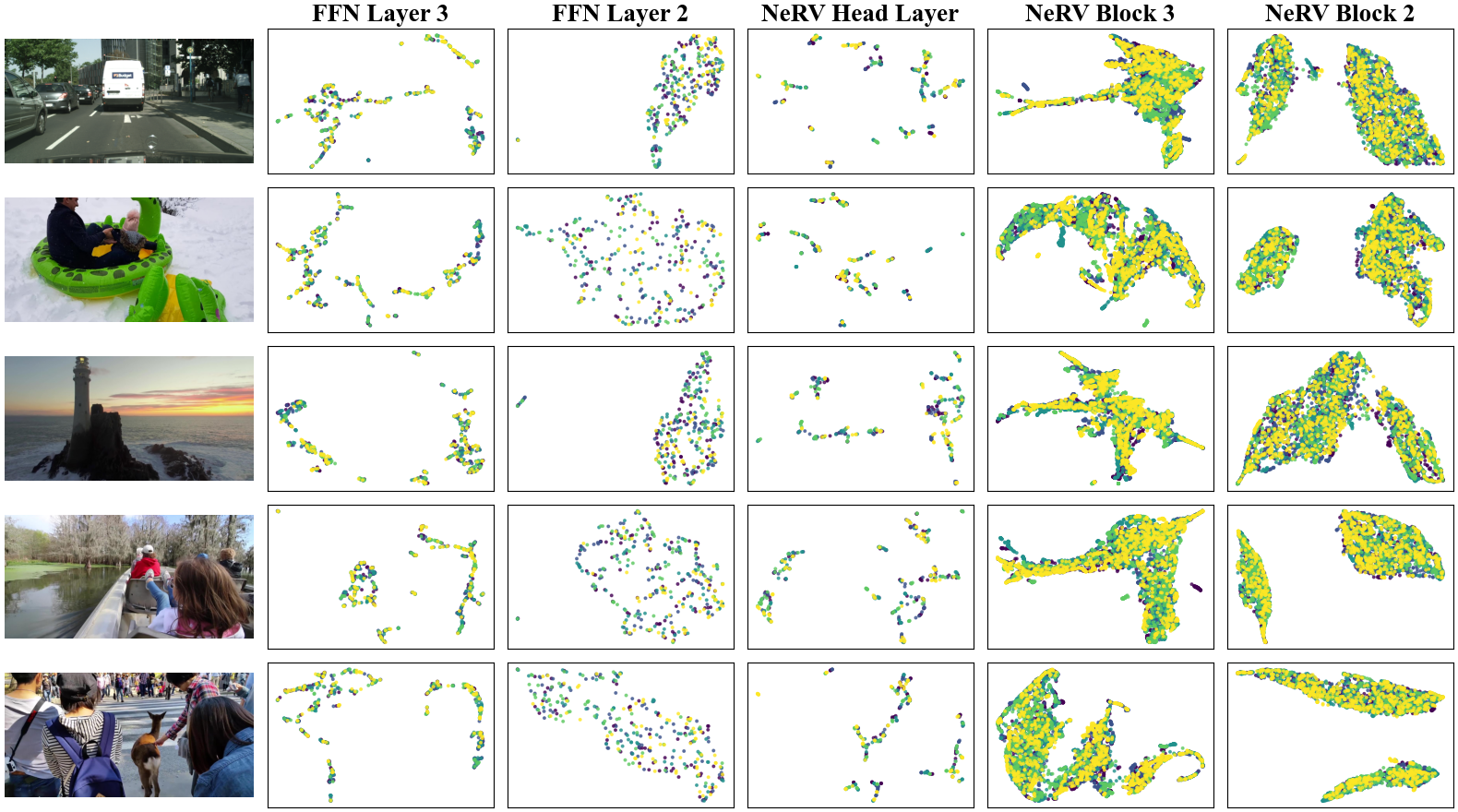}
  \caption{\textbf{Layerwise UMAP for INRs trained with different seeds}. We show results of Gabor filter based clustering neurons in each layer of MLP- and CNN-based models with different seeds. As seen in Figure \ref{fig:inr_umap}, each cluster has neurons belonging to different seeds, indicating that models of different seeds learn a set of neuron ``types''.}
  \label{fig:supp_umaps_for_seeds}
\end{figure*}

In Figure \ref{fig:supp_umaps_for_seeds}, we provide a supplement to Figure \ref{fig:inr_umap} by clustering neurons from models with five different seeds, for layers of both MLP-based and CNN-based INRs on frames from five videos.

\begin{figure*}
  \centering
   \includegraphics[width=\linewidth]{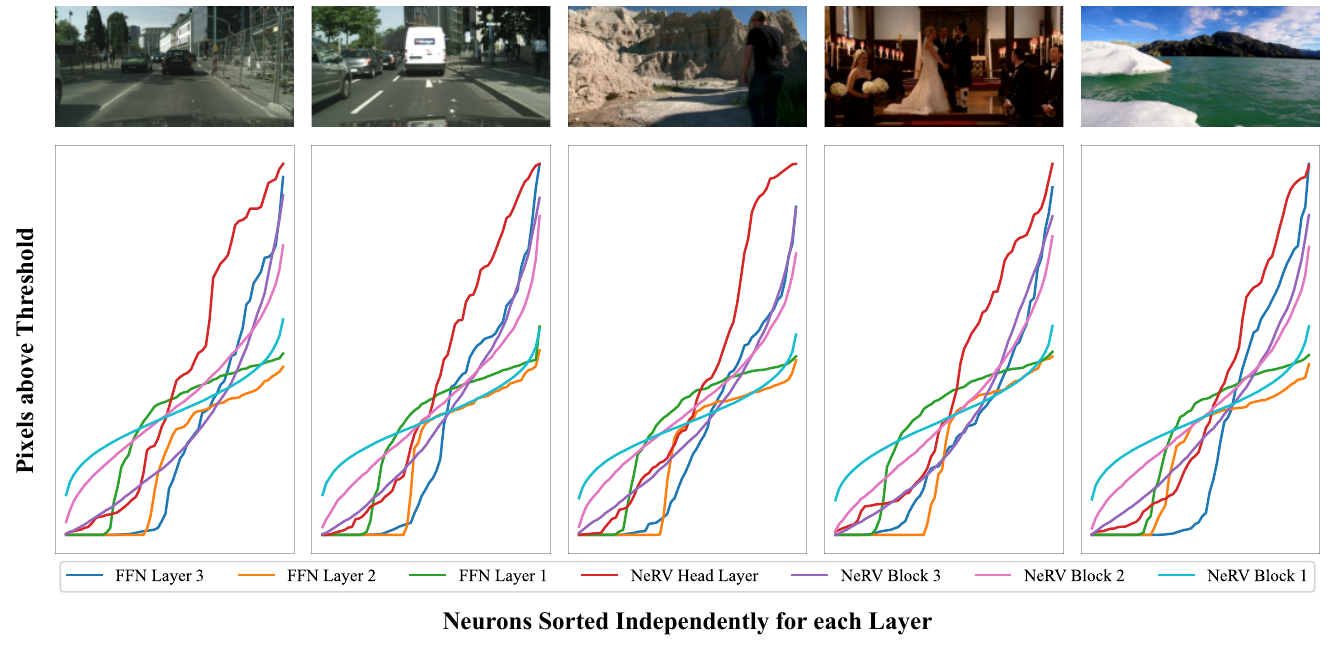}
  \caption{\textbf{Pixels per neuron.} We supplement Figure~\ref{fig:pixels_per_neuron} by plotting the pixels activated per neuron for frames from 5 additional videos. These results reveal similar trends as seen in Figure~\ref{fig:pixels_per_neuron}.}
  \label{fig:supp_pixels_per_neuron}
\end{figure*}

\subsection{Grouping Contributions}

\begin{figure*}
  \centering
   \includegraphics[width=\linewidth]{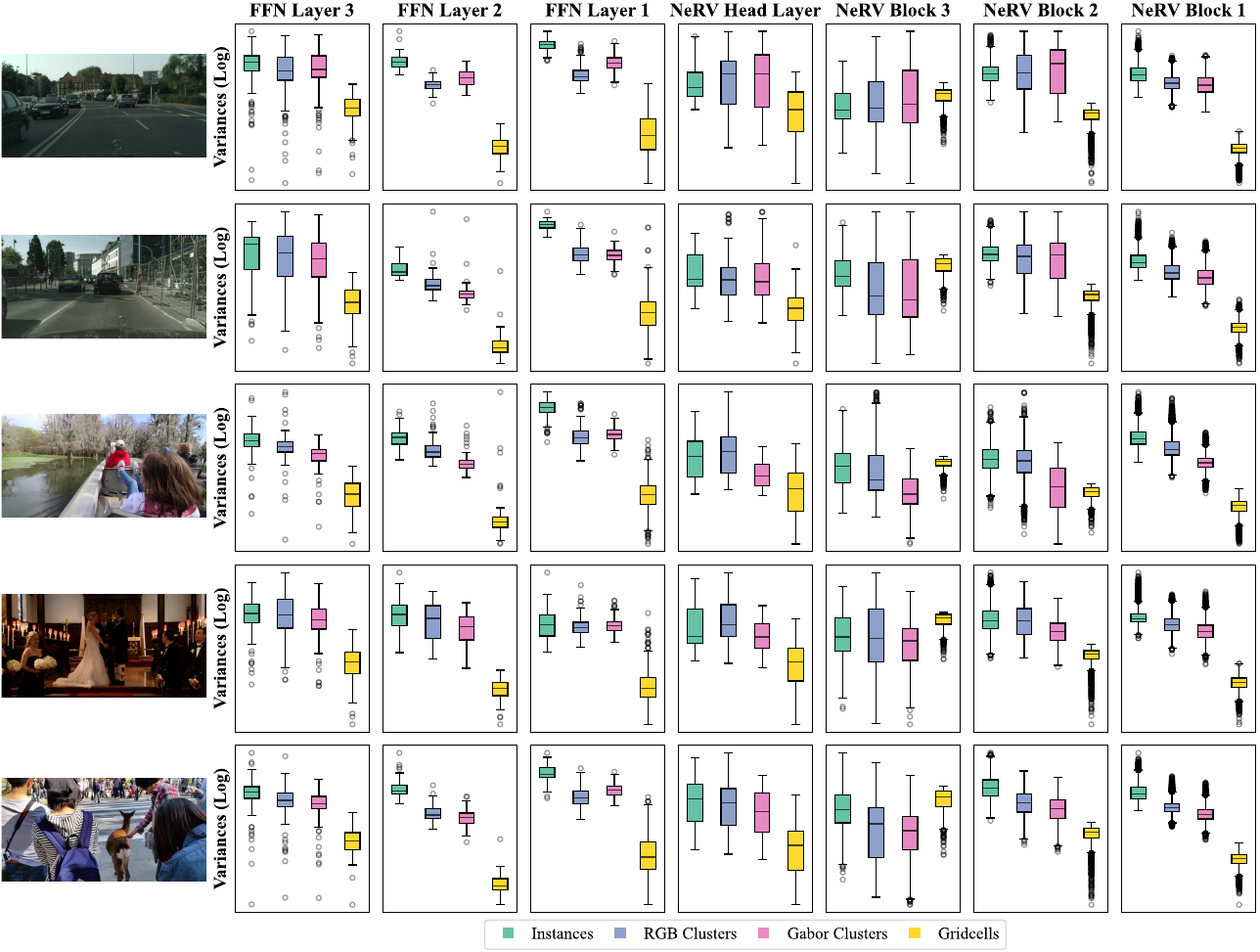}
  \caption{\textbf{Grouping contributions.} Similar to Figure \ref{fig:variance_of_patch_deltas}, we observe higher variances in neuron contribution differences when using instances, RGB clusters, and Gabor clusters in contrast to space clusters (gridcells). These observations lend support to our hypothesis that INRs prefer low-level object semantics while demonstrating a tendency to disregard space.}
  \label{fig:supp_variance_of_patch_deltas}
\end{figure*}

In Figure~\ref{fig:supp_variance_of_patch_deltas}, we provide a supplement to Figure~\ref{fig:variance_of_patch_deltas} by showing results for frames from 5 additional videos.
As expected, the major trends hold. 
We see far higher neuron contribution difference variances when using the instances, RGB clusters, and Gabor clusters, compared to space clusters (gridcells).
We see that in general Gabor and RGB clusters have equal explanatory power for neuron contributions compared to instances, reinforcing our hypothesis that these networks have low-level, rather than high-level, object semantics. 
Interestingly, the trends are consistent across all layers, except for the trends for space are weaker with the middle layers of NeRV than with the FFN, perhaps due to the spatial bias from NeRV's convolutional kernels.

\subsection{Representation is Distributed}

We extend the results from Section~\ref{subsec:representation_distributed} for frames from some additional videos.

Figure~\ref{fig:supp_pixels_per_neuron} reveals the same trends as Figure~\ref{fig:pixels_per_neuron}, for frames from 5 additional videos.
This is consistent whether the frame is street or open domain, and whether it is dim or more brightly lit.

\begin{figure*}
  \centering
   \includegraphics[width=\linewidth]{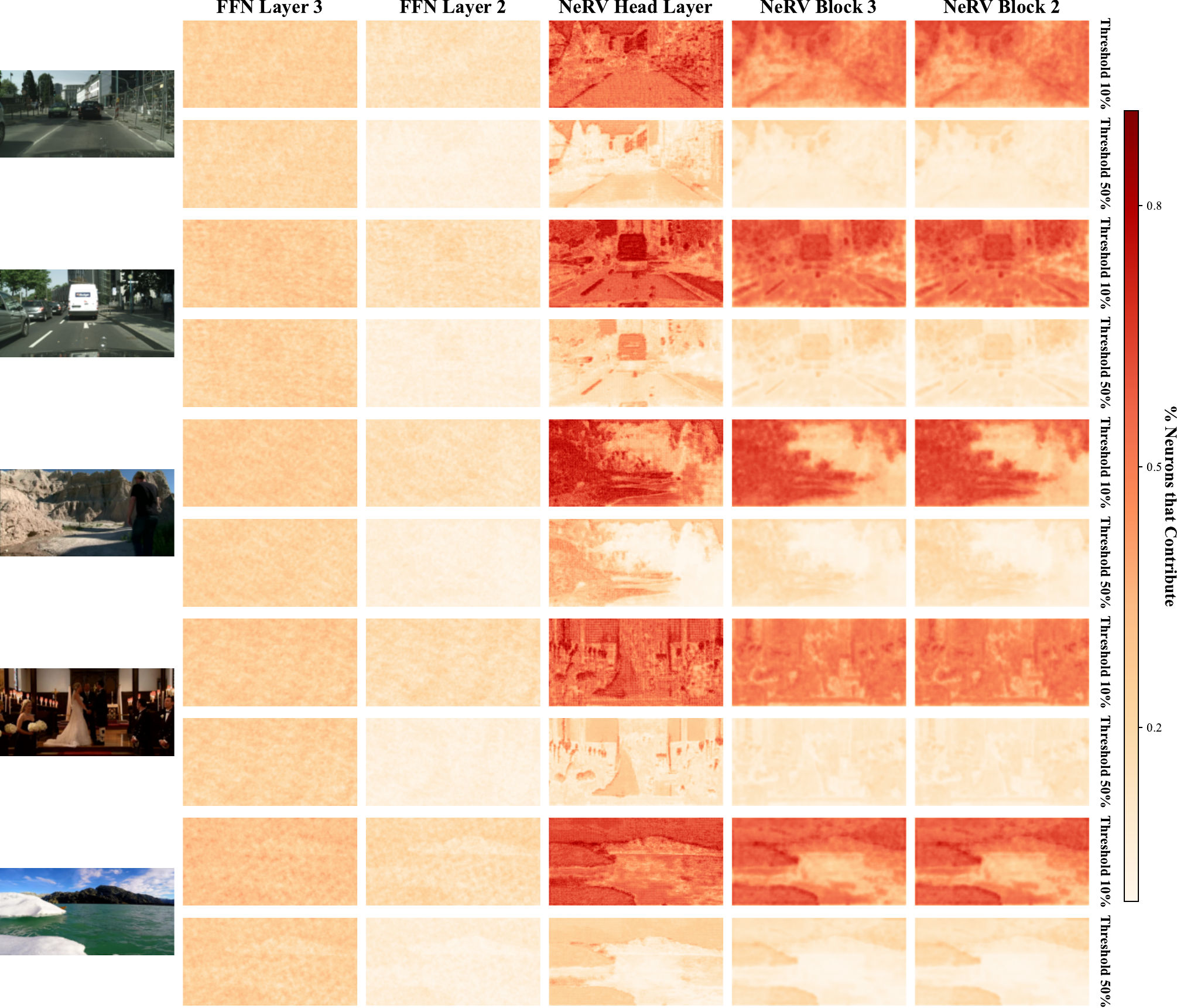}
  \caption{\textbf{Neurons per pixel.} We show the percentage of neurons in each layer, that represent significant portions of each pixel, at two different thresholds. This figure reveals properties consistent with Figure \ref{fig:neurons_per_pixel} for 5 additional video frames.}
  \label{fig:supp_neurons_per_pixel}
\end{figure*}

\begin{figure*}
  \centering
   \includegraphics[width=\linewidth]{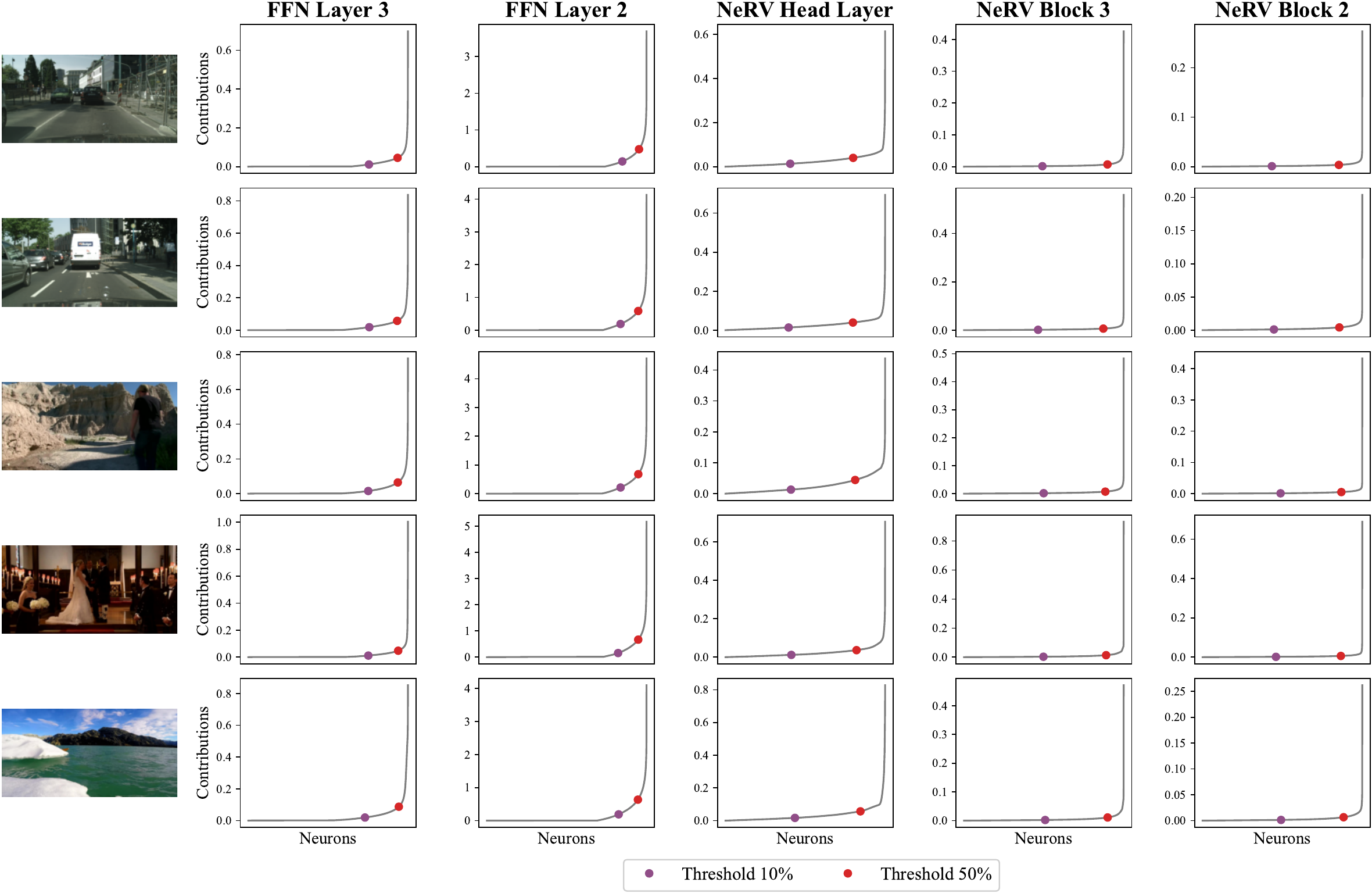}
  \caption{\textbf{Distribution of Neuron Contributions.} We plot the distribution over all neuron contributions in each layer. This shows that a large fraction of neuron contributions tend to be relatively smaller in magnitude. The contribution values corresponding to the $10^\text{th}$ and $50^\text{th}$ percentile of each distribution are used in selecting thresholds for Figures \ref{fig:neurons_per_pixel} and \ref{fig:supp_neurons_per_pixel}.}
  \label{fig:distribution_of_contributions}
\end{figure*}

\begin{figure*}
  \centering
   \includegraphics[width=\linewidth]{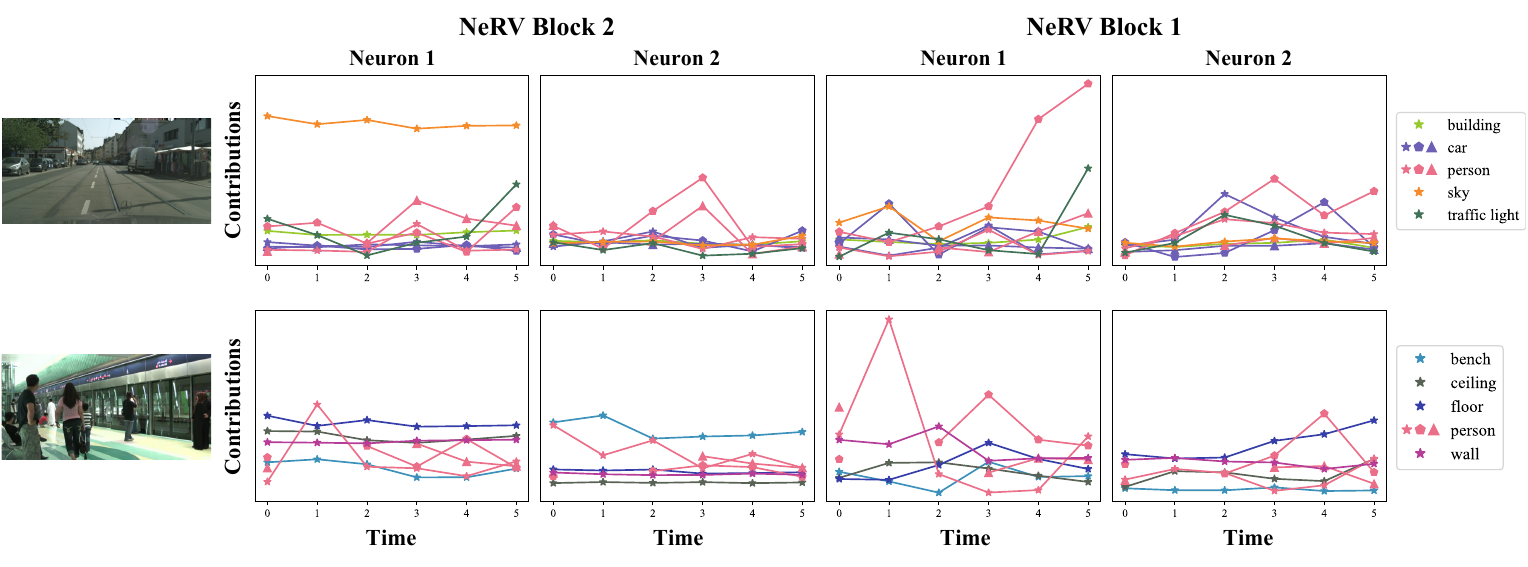}
  \caption{\textbf{Neuron contributions to things and stuff} for the first two blocks of NeRV. See Figure~\ref{fig:kernel_to_instance_contribs} for other blocks/layers.}
  \label{fig:supp_kernel_to_instance_contribs}
\end{figure*}

The same trends from the video in Figure~\ref{fig:neurons_per_pixel} are consistent for 5 additional video frames in Figure~\ref{fig:supp_neurons_per_pixel}.
Pixels for the FFN layers tend to be represented by similar amounts of neurons, whereas for the NeRV, these numbers vary widely.
Some neurons are represented by many pixels, others by very few (relatively).
The contrast is sharper in general for last layers (NeRV head layer, FFN layer 3).

Figure \ref{fig:distribution_of_contributions} shows how a large portion of the raw contributions are relatively smaller in magnitude for each layer. The thresholds for Figure \ref{fig:supp_neurons_per_pixel} are selected using the contribution value at the $10^\text{th}$ and $50^\text{th}$ percentile of these curves.

\subsection{Objects and Categories}

We offer results for the first two NeRV blocks corresponding to Figure~\ref{fig:kernel_to_instance_contribs} in Figure~\ref{fig:supp_kernel_to_instance_contribs}, reserved for this appendix due to space constraints.
Overall, the representations for objects are relatively constant over time for Block 2, with some notable exceptions, such as the representation of one of the persons for one of the four sampled neurons, which increases dramatically at the final frame of the video.
We also note that Block 1 seems less structured than the other blocks.
Specifically, whereas certain object types might dominate for the other blocks, in Block 1 the contributions are more evenly distributed across objects overall. 

\end{document}